\documentclass[letterpaper, 10 pt, conference]{ieeeconf}  
\usepackage{booktabs}
\usepackage{graphicx}
\usepackage{subcaption}
\usepackage{subcaption}
\usepackage{epstopdf}
\usepackage[utf8]{inputenc}
\usepackage{float}
\usepackage{cite}
\usepackage{placeins}  
\usepackage{amsmath} 
\usepackage{amssymb} 
\usepackage{amsmath}
\usepackage{subcaption}
\usepackage{booktabs}
\usepackage{multirow}
\usepackage{makecell}

\IEEEoverridecommandlockouts                              

\title{\LARGE \bf
Gaussian Sequences with Multi-Scale Dynamics for 4D Reconstruction from Monocular Casual Videos}

\author{
Can Li$^{1, \#}$, \thanks{$^{1}$ Nankai University, Tianjin, China. }\thanks{$^{\#}$ Work was done during the internship at Rightly Robotics. } 
Jie Gu$^{2}$, \thanks{$^{2}$ Rightly Robotics, Hangzhou, China.} 
Jingmin Chen$^{2}$,
Fangzhou Qiu$^{2}$,
and Lei Sun$^{1}$
}

\begin{document}

\maketitle
\thispagestyle{empty}
\pagestyle{empty}

\begin{abstract}

Understanding dynamic scenes from casual videos is critical for scalable robot learning, yet four-dimensional (4D) reconstruction under strictly monocular settings remains highly ill-posed. To address this challenge, our key insight is that real-world dynamics exhibits a multi-scale regularity from object to particle level.
To this end, we design the multi-scale dynamics mechanism that factorizes complex motion fields.  
Within this formulation, we propose Gaussian sequences with multi-scale dynamics, a novel representation for dynamic 3D Gaussians derived through compositions of multi-level motion. This layered structure substantially alleviates ambiguity of reconstruction and promotes physically plausible dynamics.
We further incorporate multi-modal priors from vision foundation models to establish complementary supervision, constraining the solution space and improving the reconstruction fidelity.
Our approach enables accurate and globally consistent 4D reconstruction from  monocular casual videos. Experiments of dynamic novel-view synthesis (NVS) on benchmark and real-world manipulation datasets demonstrate considerable improvements over existing methods. 

\end{abstract}

\section{INTRODUCTION}

Embodied AI aims to endow robotic agents with the ability to perceive, reason, and act in the physical world. Achieving this goal critically depends on large-scale training data that capture both real environments and robot actions~\cite{o2024open, khazatsky2024droid, chi2025diffusion, Zhao-RSS-23}. Unfortunately, real environments in such training data are typically captured as videos, providing no insight into the underlying three-dimensional space or the scene dynamics behind the observations. This gap motivates reconstructing four-dimensional (4D) representations directly from monocular videos. By bridging this gap, 4D reconstruction allows robots to effectively exploit visual data to infer spatio-temporal geometry and predictive dynamics or world models~\cite{ai2025review}, ultimately contributing to scalable robot learning. Therefore, it is not only technically compelling but also strategically essential in the era of embodied intelligence.

\begin{figure}[t]
    \centering
    \begin{subfigure}[b]{0.48\textwidth}
        \centering
        \includegraphics[width=\textwidth]{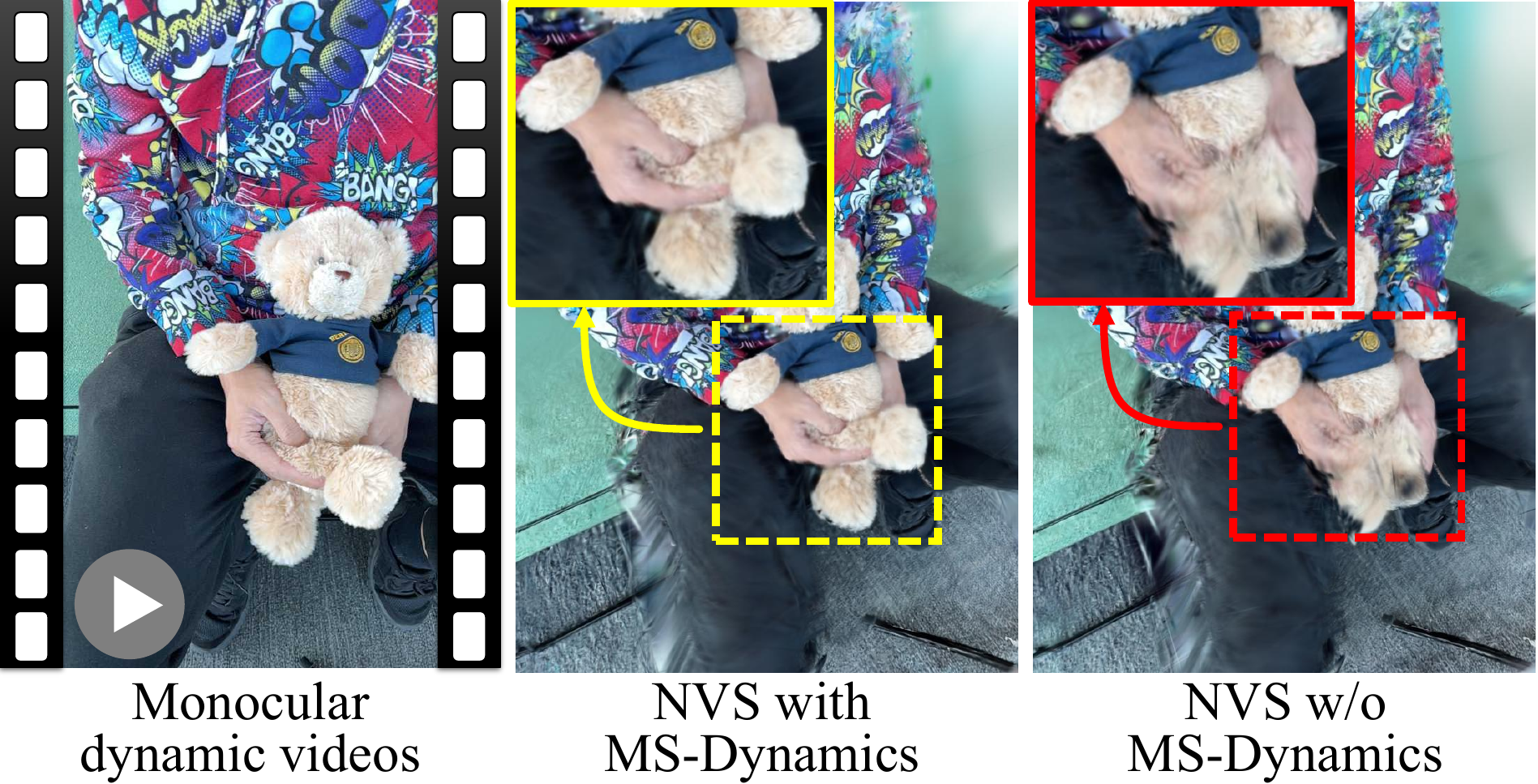}  
        \caption{Monocular 4D Gaussian reconstruction of HOI.}
    \end{subfigure}
    \\[1mm] 
    \begin{subfigure}[b]{0.48\textwidth}
        \centering
        \includegraphics[width=\textwidth]{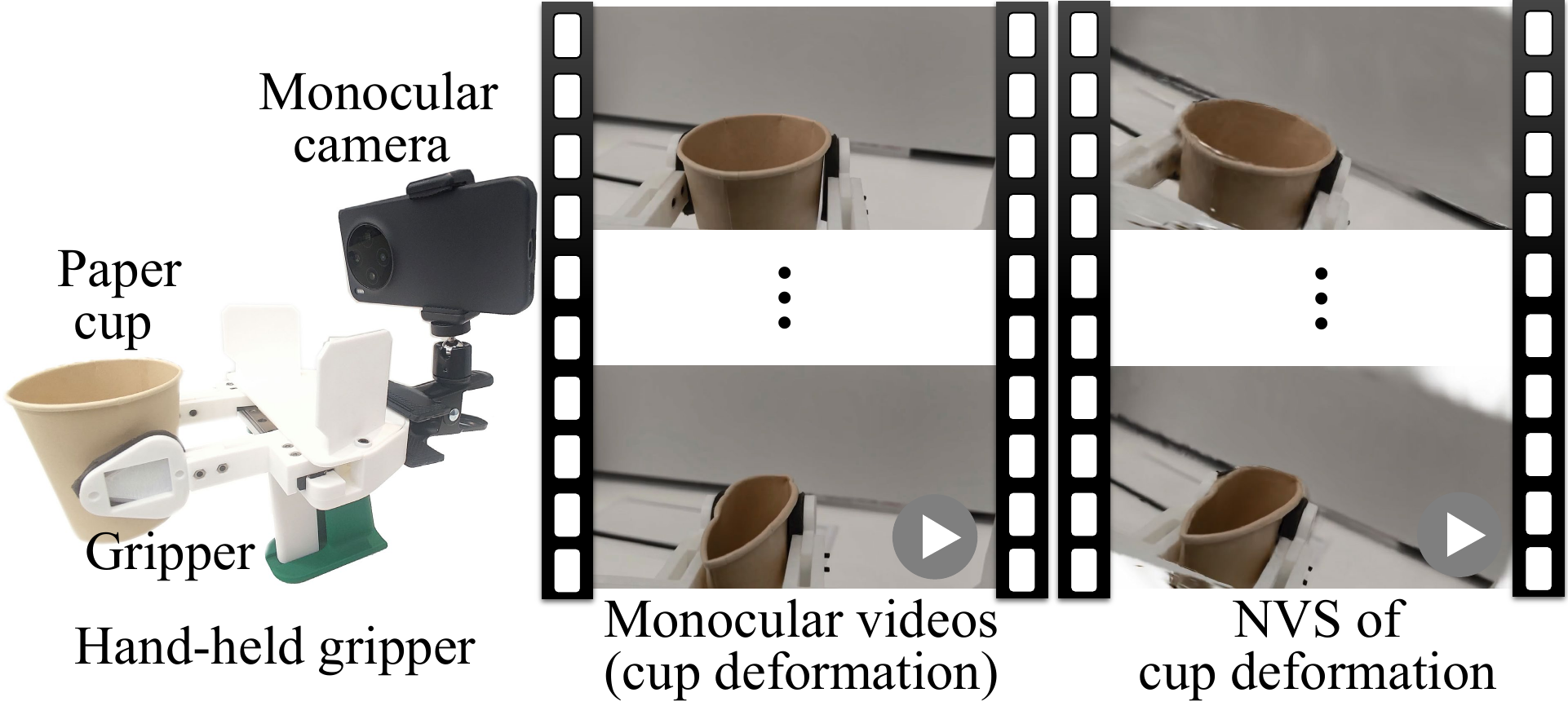}  
        \caption{Application to hand-held data collection.}
    \end{subfigure}
    \caption{\textbf{(a)} From a casually captured monocular video with complex hand–object interactions (left), our MS-Dynamics models multi-scale dynamics to drive 4D Gaussians, producing a temporally-coherent Gaussian sequence that synthesizes novel views with fine hand details (middle). Without MS-Dynamics, the 4D reconstruction is blurry and lacks structural fidelity (right). \textbf{(b)} An application to hand-held data collection (left): starting from a video demonstration of cup deformation (middle), our method generates novel-view demonstrations (right).}
    \label{fig:demo}
\end{figure}

Reliable 4D reconstruction of dynamic scenes remains highly challenging, especially under in-the-wild monocular conditions. Although static spatial reconstruction has been greatly advanced by 3D Gaussian Splatting (3DGS)~\cite{kerbl20233d}, monocular 4D reconstruction is fundamentally ill-posed: the solution space is vast and supervision is extremely limited, often resulting in fragile and unstable reconstructions. The difficulty arises from the nature of monocular video data, \emph{e.g.}, sparse observations, limited parallax, depth ambiguity, and inherent scene dynamics. Moreover, dynamic scenes typically involve nonlinear deformations, frequent occlusions, and complex object interactions, further compounding the challenges.


A natural strategy for 4D reconstruction is to extend 3DGS by implicitly learning the temporal evolution of Gaussians. Warping Gaussians over time using implicit deformation fields~\cite{yang2024deformable, wu20244d} shows promise, but these approaches often oversmooth dynamic details and are primarily evaluated in controlled settings. Furthermore, their “monocular” setups typically rely on camera orbits around dynamic objects that approximate multi-view capture, leaving their effectiveness in real-world monocular videos unclear.


Explicitly modeling the temporal evolution of Gaussians offers a promising alternative~\cite{lei2025mosca, wang2025shape}. MoSca~\cite{lei2025mosca} uses a large number of $SE(3)$-deformation nodes to model Gaussian motion, achieving high expressiveness but introducing excessive degrees of freedom that are prone to overfitting. Shape-of-motion~\cite{wang2025shape} instead employs a low-dimensional motion field that exploits motion smoothness, improving stability but lacking the flexibility to model fine-grained deformations. These limitations highlight the need for a representation that is expressive enough to capture fine-grained dynamics yet sufficiently regularized to avoid overfitting under monocular supervision. 


A central insight of this work is that the ill-posedness of monocular 4D reconstruction can be substantially alleviated by leveraging the inherent multi-scale regularities of real-world dynamics. Specifically, motion in natural scenes is not arbitrary. It is structured, unfolding from global object-level trajectories down to fine local deformations. These multi-scale patterns serve as implicit physical constraints that all dynamic scenes must satisfy. 


Building on this insight, we design a multi-scale dynamics (MS-Dynamics) mechanism that explicitly models this layered motion structure. By factorizing dynamics across object-level motion, primitive-level transformations, and fine-grained local deformations, MS-Dynamics reduces ambiguity and guides the optimization toward physically plausible solutions. This structured factorization injects strong inductive bias, enabling the representation to remain highly expressive while simultaneously constraining the solution space—striking precisely the balance required for robust monocular 4D reconstruction.


We further observe that priors from vision foundation models, whose capabilities have grown significantly in recent years~\cite{yang2023track, Yang2024DepthAU, piccinelli2024unidepth, Doersch2023TAPIRTA},  provide valuable complementary supervision beyond raw RGB values, alleviating the inherent sparsity of supervision in monocular videos. Although the generated priors may not be perfectly accurate and mild noise is unavoidable in practice, their combination effectively constrains the solution space and improves the fidelity of dynamic reconstruction. 

Consequently, we propose Gaussian Sequences with MS-Dynamics, where dynamic Gaussians are derived with multi-scale motion and supervised by multi-modal prior signals. This formulation enables reliable reconstruction of Gaussian sequences from monocular videos and supports high-quality dynamic novel-view synthesis (NVS), as presented in Figure~\ref{fig:demo}.

In summary, our key contributions are as follows:
\begin{itemize}
\item We design the MS-Dynamics mechanism that represents complex dynamics through multi-scale motion fields. This factorization provides strong inductive bias and effectively reduces motion ambiguity.
\item We propose Gaussian sequences with MS-Dynamics for robust 4D reconstruction from monocular casual videos, guided by complementary multi-modal priors. This formulation strongly alleviates the high ill-posedness of 4D monocular reconstruction, improving reconstruction stability and consistency in strictly-monocular settings. 
\item We provide custom datasets with rigid, articulated, and deformable objects, and extensively evaluate our approach on both benchmark and custom data, showing considerable gains in dynamic NVS.
\end{itemize}

\section{related works}
\subsection{3D Reconstruction}

Recent years have witnessed rapid progress in NVS for static scenes. Neural Radiance Fields (NeRF) represent scenes using coordinate-based MLPs~\cite{mildenhall2021nerf}, achieving impressive photo-realistic results but suffering from slow rendering. 3DGS~\cite{kerbl20233d}  offers a more efficient alternative by representing scenes as a set of anisotropic 3D Gaussians, achieving real-time rendering without sacrificing quality. Building on this, many recent works have enhanced 3DGS and extended its applications to robotics, spanning SLAM~\cite{Hong2024LIV}, manipulation~\cite{Lu20243DGSCD3G, YangS-RSS-25}, and teleoperation~\cite{Lee2025Human}.

However, existing methods focus mainly on static scenes. Our work addresses this gap by a dynamic Gaussian-based representation tailored for monocular 4D reconstruction.

\subsection{Dynamic Reconstruction}

The goal of dynamic reconstruction is to recover the geometry, appearance, and motion that evolve over time. Although impressive results have been achieved using multi-view synchronized cameras~\cite{fridovich2023k, luiten2024dynamic}, such settings are often impractical in real-world applications. As a result, monocular reconstruction has attracted increasing attention for its simplicity and applicability~\cite{lei2025mosca, qingming2025modgs, stearns2024dynamic, wang2025shape}.

NeRF-based methods learn deformation fields jointly with radiance fields~\cite{park2021nerfies, park2021hypernerf}. These deformation fields parameterized by MLPs benefit from continuity and compactness but struggle to capture fine-grained or highly non-rigid motion due to their over-smoothness.

Several Gaussian splatting-based methods adopt a similar latent deformation-based strategy~\cite{wu20244d, yang2024deformable}. However, despite the rendering efficiency of 3DGS, the combination with implicit deformation fields still inherits limitations in motion fidelity and computational cost. To address these issues, some works replace implicit fields with explicit motion modeling~\cite{wang2025shape, lei2025mosca, park2025splinegs}. For example, Shape-of-motion~\cite{wang2025shape} utilizes shared low-dimension motion bases, and SplineGS~\cite{park2025splinegs} leverages motion-adaptive Hermite splines and joint optimization to enable NVS from monocular videos.

These advances highlight the importance of dynamic representations, yet notable challenges remain: low-rank models struggle to capture fine-grained motion, whereas high-dimensional formulations easily overfit in monocular settings. These limitations motivate us to pursue a balanced representation with multi-scale dynamics.

\section{Methods}

\begin{figure*}[t]
\centering
\includegraphics[width=0.96\textwidth]{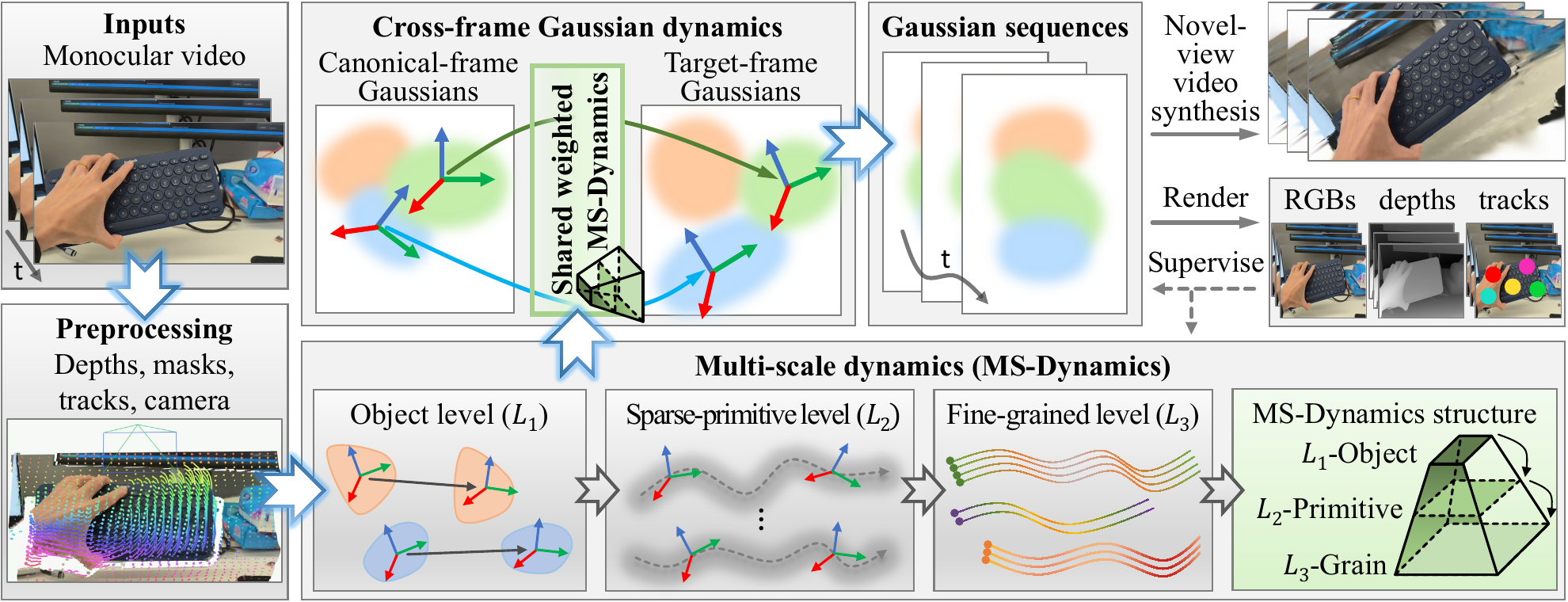} 
\caption{Overview of \textbf{Gaussian Sequences with MS-Dynamics} for 4D monocular reconstruction. The pipeline first preprocesses monocular videos to obtain depths, masks, point tracks, and camera parameters. Our MS-Dynamics performs multi-scale factorization from object ($L_1$), through sparse-primitive ($L_2$), to fine-grained level ($L_3$), capturing both global motion and local detailed deformation. Cross-frame Gaussian dynamics from canonical to target frame is modeled by shared weighted MS-Dynamics, constructing globally consistent Gaussian sequences. Both Gaussian sequences and MS-Dynamics are supervised by the aggregation of multi-modal signals (such as RGBs, depths, and tracks), which provides complementary cues for globally consistent optimization. The resulting Gaussian sequences enable high-quality dynamic NVS.}
\label{fig-pipeline}
\end{figure*}

As shown in Figure~\ref{fig-pipeline}, we propose Gaussian Sequences with MS-Dynamics for 4D reconstruction from monocular casual videos, yielding a dynamic Gaussian representation that enables photo-realistic NVS at arbitrary times. First of all, section \ref{subsec-preliminaries}  reviews the fundamentals of 3DGS. Section \ref{subsec-Formulation} provides the problem formulation, casting 4D reconstruction as the optimization of time-varying Gaussians. In section \ref{subsec-GaussianSequences}, we thoroughly explain our framework of Gaussian sequences with MS-Dynamics. Finally, section \ref{subsec-loss} describes the well-designed loss functions for supervision.

\subsection{Preliminaries}
\label{subsec-preliminaries}

3DGS~\cite{kerbl20233d} models a static scene through a collection of anisotropic 3D Gaussian primitives, allowing efficient photo-realistic rendering. Each Gaussian is defined as: 
\begin{equation}
\mathcal{G}(\boldsymbol{\mu}, \mathbf{\Sigma}, \alpha, \mathbf{c})
\end{equation}
with a mean $\boldsymbol{\mu} \in \mathbb{R}^3$, a covariance matrix $\mathbf{\Sigma} \in \mathbb{R}^{3 \times 3}$, an opacity $\alpha \in \mathbb{R}^+$, and a view-dependent color $\mathbf{c} \in \mathbb{R}^{3(l+1)^2}$ parameterized via spherical harmonics of degree $l$. Moreover, the covariance matrix $\mathbf{\Sigma}$ can be decomposed into a rotation and a scale component as:
\begin{equation}
\mathbf{\Sigma} = \mathbf{R}_{\Sigma} \, \mathbf{S}\mathbf{S}^\top \, \mathbf{R}_{\Sigma}^\top,
\end{equation}
where $\mathbf{R}_{\Sigma} \in SO(3)$ is a rotation matrix and $\mathbf{S} \in \mathbb{R}_+^3$ determines the scales.

To render the scene from a camera with parameters $\mathbb {\theta}$, each 3D Gaussian is projected to 2D as: 
\begin{equation}
\boldsymbol{\mu}' = \Pi(\boldsymbol{\mu}; \boldsymbol{\theta})  \in \mathbb{R}^{2}, \
\mathbf{\Sigma}' = \Pi(\mathbf{\Sigma}; \boldsymbol{\theta}) \in \mathbb{R}^{2 \times 2}, 
\end{equation}
where $\Pi(\cdot)$ denotes the projection function. The 2D Gaussians are then sorted by depth and composited using alpha blending:
\begin{equation}
C(\mathbf{p}) = \sum_{i=1}^N \mathbf{c}_i \, \sigma_i(\mathbf{p}) \left( \prod_{j=1}^{i-1} \left(1 - \sigma_j(\mathbf{p}) \right) \right),
\end{equation}
\begin{equation}
\sigma_i(\mathbf{p}) = \alpha_i \exp \left( -\frac{1}{2} (\mathbf{p} - \boldsymbol{\mu}'_i)^\top (\mathbf{\Sigma}'_i)^{-1} (\mathbf{p} - \boldsymbol{\mu}'_i) \right),
\end{equation}
where $\mathbf{p} \in \mathbb{R}^2$ is the 2D pixel coordinate, and $N$ is the number of Gaussians intersecting the ray through $\mathbf{p}$.

\subsection{Problem Formulation}
\label{subsec-Formulation}
To effectively handle dynamic conditions, we need to represent the evolving state and appearance of dynamic objects over time. Thus, we formulate 4D reconstruction as estimating a time-varying and temporally consistent Gaussian field $\mathcal{G}_t$:
\begin{equation}
\mathcal{G}_t := \left\{ \mathcal{G}_t^i \right\}_{i=1}^N = \left\{\mathcal{T}_t^i\left(\mathcal{G}_0^i\right) \right\}_{i=1}^N=\left\{ \mathbf{T}_t^i \odot \mathcal{G}_0^i\right\}_{i=1}^N, 
\label{eq-formulation-main-1}
\end{equation}
where $t$ denotes the time index and $\mathcal{G}_t^i$ is a dynamic Gaussian derived by applying a transformation operator $\mathcal{T}_t^i$ to the canonical Gaussian $\mathcal{G}_0^i$. 
If we parameterize $\mathcal{G}_0^i=\left(\boldsymbol{\mu}_0^i, \mathbf{\Sigma}_0^i, \mathbf{c}^i, \alpha^i\right)$ and $\mathbf{T}_t^i=\left[\mathbf{R}_t^i, \mathbf{t}_t^i\right]$, the operator $\mathcal{T}_t^i$ acts on $\mathcal{G}_0^i$ as:
\begin{equation}
\boldsymbol{\mu}_t^i=\mathbf{R}_t^i \boldsymbol{\mu}_0^i+\mathbf{t}_t^i, \quad \mathbf{\Sigma}_t^i=\mathbf{R}_t^i \mathbf{\Sigma}_0^i\left(\mathbf{R}_t^i\right)^{\top}.
\label{eq-formulation-detail}
\end{equation}
Here, we assume that only the position and orientation of each Gaussian varies over time, while scale, color, and opacity remain static.

Therefore, the 4D Gaussian reconstruction problem is formulated as the following joint optimization:
\begin{equation}
    \min_{\{\mathcal{G}_0^i\}_{N_G},\,\{\mathbf{T}_t^i\}_{N_T}}
    \;
    \sum_{t=1}^{T}
    \mathcal{L}_{\mathrm{render}}\bigg(\mathcal{R} \Big( \left\{\mathbf{T}_t^i \odot \mathcal{G}_0^i\right\}_{i=1}^N \Big),\, \boldsymbol{O}_{\text{mono}} \bigg),
\label{eq-optimization}
\end{equation}
where $\{\mathcal{G}_0^i\}_{i=1}^{N_G}$ and $\{\mathbf{T}_t^i\}_{i=1}^{N_T},\forall t \in {1, \dots, T}$ 
are the variables to be optimized, $\mathcal{R}$ is a render operator, $\boldsymbol{O}_{\text{mono}}$ denotes the monocular observation, and $\mathcal{L}_{\mathrm{render}}$ is the rendering loss. 

Although Eq. (\ref{eq-optimization}) implies estimating an independent motion for each Gaussian, doing so would lead to an intractably high-dimensional optimization. Instead, we learn a shared low-rank, multi-scale motion field for dynamic Gaussians. Below, we describe how this is constructed.

\subsection{Gaussian Sequences with MS-Dynamics}
\label{subsec-GaussianSequences}
Dynamic scenes captured from a monocular video often exhibit object interactions, non-rigid deformations, and spatially heterogeneous motion. Modeling such complex motion directly at per Gaussian is highly under-constrained: each Gaussian moves in 3D space, but its motion is only weakly observed from monocular frames. This leads to severe ambiguity, temporal drift, and overfitting when learning fully independent transformations for every Gaussian. 

To address these challenges, our key insight is that real-world dynamic scenes contain strong multi-scale and low-rank regularities. Object-level motion tends to be globally coherent; within deformable objects, motion can be decomposed into a small number of principal deformation patterns; and at the finest scale, individual particles vary smoothly around these shared bases. 
This motivates a multi-scale dynamics modeling approach that shares motion structures across all Gaussians while allowing each Gaussian to make fine-scale adjustments through its own learnable weights.

\subsubsection{Overview of MS-Dynamics} 

Figure~\ref{fig-pipeline} illustrates the proposed method of Gaussian sequences with MS-Dynamics. To model complex high-dimension dynamics, we propose shared weighted MS-Dynamics to explicitly factorize the per-Gaussian transformation $\mathbf{T}_t^i$ in Eq. (\ref{eq-formulation-main-1}), capturing non-rigid motion with low-rank but multi-scale motion field. The general transformation $\mathbf{T}_t^i$ is first formulated hierarchically in three layers:
\begin{equation}
\mathbf{T}_t^i = \prod_{l=1}^{3} \mathbf{T}_t^{i,(L_l)},
\label{eq-hierachy}
\end{equation}
representing the composition of relative transformations in a coarse-to-fine manner, with layers corresponding to object-level ($L_1$), sparse-primitive-level ($L_2$), and fine-grained-level ($L_3$) motion.

In each layer $L_l$, $\mathbf{T}_t^{i,(L_l)}$ itself is a weighted combination of the shared principal motion patterns. To simplify notation in $SE(3)$, we approximate the transformations of $\mathbf{T}_t^{i,(L_l)}$ as:
\begin{equation}
\mathbf{T}_t^{i,(L_l)} \approx \sum_{k=1}^{K_l} w_t^{i,(L_l, k)} \mathbf{P}_t^{(L_l, k)}, \quad l=\text{1, 2, 3,}
\label{eq-weighted combination}
\end{equation} 
where $K_l$ is the number of principal motion patterns in the layer $L_l$, $\mathbf{P}_t^{(L_l, k)} \in SE(3)$ is the $k$-th shared motion pattern, and $w_t^{i,(L_l, k)}$ are the learnable weights for Gaussian $i$. 

This design allows for sharing global dynamics across objects, low-rank deformation patterns inside each object, and flexible local corrections at grain level.

\subsubsection{Object Level ($L_1$)}
General objects, even deformable ones, often exhibit motion that is coherent to some extent: parts of the object tend to move in a correlated manner, even when the overall shape deforms. 
We derive object-level motion patterns by clustering 3D point tracks into object groups and estimating their dominant rigid (or approximately coherent) transformations. Then, dynamic Gaussians share several object-level motion patterns. The reason for sharing is that different objects, especially during object interactions, often exhibit similar motion modes. 

Given preprocessed 3D point tracks $\{\boldsymbol{x}_t^j\}_{j=1}^N$ and object segmentation masks, we first cluster the tracks into object-level groups. 
For each object-level cluster $\mathcal{C}^{(L_1, k)}, k=1,...,K_{\text{object}}$, we compute a dominant rigid transformation $\mathbf{P}_t^{(L_1, k)}$ of each cluster via weighted Procrustes alignment:
\begin{equation}
\mathbf{P}_t^{(L_1, k)} = \arg\min_{\mathbf{R}, \mathbf{t}} \sum_{\boldsymbol{x}^j \in \mathcal{C}^{(L_1, k)}} w_j \left\| \mathbf{R} \boldsymbol{x}_0^j + \mathbf{t} - \boldsymbol{x}_t^j \right\|^2,
\label{eq-wPA}
\end{equation}
where $\boldsymbol{x}_0^j$ and $\boldsymbol{x}_t^j$ are the corresponding point sets of canonical and target frames, respectively, and $w_j$ encodes the confidence of each point track.  

With $K_{\text{object}}$ shared motion patterns $\mathbf{P}_t^{(L_1, k)}$, canonical-frame Gaussians can be transformed to target-frame ones using Eq. (\ref{eq-weighted combination}) at the object level, where each Gaussian is assigned soft learning weights $w_t^{i,(L_1, k)}$ based on its spatial proximity to the corresponding object cluster. 

The object-level motion field is too coarse to be directly applied to represent Gaussian dynamics. However, capturing this object level provides a stable coarse estimate that anchors the subsequent layers of the multi-scale motion hierarchy.  

\subsubsection{Sparse-primitive Level ($L_2$)}

Non-rigid motion exists within deformable objects. Directly learning per-Gaussian motion is unstable; instead, deformation usually lies in a low-rank subspace.
Inspired by the low-rank property, at this level, we capture object-internal deformation using a set of sparse motion primitives, each representing a principal mode in non-rigid deformation. With shared weighted sparse primitives, Gaussian deformation can be derived. 

Within each object region, we first group the tracked 3D trajectories according to their movement directions. A weighted Procrustes alignment is then applied to each trajectory cluster, similar to Eq. (\ref{eq-wPA}), yielding $K_{\text{primitive}}$ motion bases per object. By assigning soft weights to these shared motion bases as in Eq. (\ref{eq-weighted combination}), the dynamic Gaussians can be driven by a blend of them, enabling expressive non-rigid motion while maintaining a compact motion representation.

Consequently, this level produces low-rank, sparse, and shared deformation modes. Note that $L_2$ deformation is defined under the coordinate reference of object levels.

\subsubsection{Fine-grained Level $(L_3)$} 

Although object-level and sparse-primitive-level motion capture the coarse and mid-level dynamics, they are insufficient to represent the subtle surface variations required for high-quality dynamic Gaussian reconstruction.

At the fine-grained level, we move toward denser and more local motion patterns, refining local deformations and correcting misalignments. Specifically, around each sparse primitive, we compute the residual motion between the primitive and the nearby point trajectories. These residuals are then clustered, and for each cluster we estimate a residual fine-grained motion modes using Eq. (\ref{eq-wPA}). $K_{\text{grain}}$ groups are borned at the $L_3$ level for every sparse primitive ($L_2$).

Therefore, these fine-grained residual modes further enhance the expressiveness of the motion field, enabling the Gaussians to faithfully capture fine-scale and highly local deformations that cannot be explained by preceding motion levels.

\subsubsection{MS-Dynamics Structure}

After obtaining the motion fields from the three levels described above, we combine them according to Eq. (\ref{eq-hierachy}) to construct the MS-Dynamics Hierarchy, as illustrated in Figure~\ref{fig-pipeline}. Note that motion across levels is defined relatively, with each level expressed in the reference frame of the previous one.

Although theoretically one could further explore a dense particle–level dynamics representation, such an approach would significantly increase the dimensionality of optimization and lead to unstable solutions. In contrast, our MS-Dynamics hierarchy achieves a desirable trade-off between low-rank structure and multi-scale expressiveness.

Each Gaussian shares the MS-Dynamics hierarchy to follow globally coherent object-level motion while simultaneously capturing local fine-grained deformations. This exploits the underlying low-rank and multi-scale structure of dynamic scenes, enabling motion primitives to be efficiently learned and reused across all Gaussians, thereby improving both computational efficiency and temporal consistency.

\subsection{Loss Functions}
\label{subsec-loss}
To alleviate the inherently ill-posed nature of monocular video reconstruction, \emph{i.e.}, the optimization in Eq. (\ref{eq-optimization}), we incorporate a set of comprehensive multi-modal supervisions from monocular observations $\boldsymbol{O}_{\text{mono}}$. These multi-modal signals are acquired from off-the-shelf visual foundation models, contributing to constrain the solution space and improve the reconstruction fidelity.

Specifically, we employ a segmentation foundation model~\cite{yang2023track} to extract dynamic object masks, a monocular depth estimator~\cite{Yang2024DepthAU, piccinelli2024unidepth} to obtain depth maps, and a track-any-points model~\cite{Doersch2023TAPIRTA} to estimate long-term dense point trajectories.

The above results are then integrated to form a comprehensive multi-modal supervision signal. Accordingly, the loss function of the optimization in Eq. (\ref{eq-optimization}) comprises a RGB-image loss $\mathcal{L}^{\mathrm{\text {rgb}}}$, a dynamic mask loss $\mathcal{L}^{\mathrm{\text {mask}}}$, a depth alignment loss $\mathcal{L}^{\mathrm{\text {depth}}}$, and a tracking loss $\mathcal{L}^{\mathrm{\text {track}}}$. In addition, we include a local rigidity loss $\mathcal{L}^{\mathrm{\text{local-rigid}}}$ to regularize the motion of dynamic Gaussians, enforcing the motion consistency among neighbors of Gaussians.

Consequently, we supervise the optimization of Eq. (\ref{eq-optimization}) with total loss $\mathcal{L}^{\text {total}}$ as:
\begin{align}
\mathcal{L}^{\text {total}} 
&= \lambda_1 \mathcal{L}^{\text {rgb}} + \lambda_2 \mathcal{L}^{\text {mask}} + \lambda_3 \mathcal{L}^{\text {depth}} \nonumber \\
&\quad + \lambda_4 \mathcal{L}^{\text {track}} + \lambda_5 \mathcal{L}^{\text {local-rigid}}.
\label{eq-loss}
\end{align}

\section{experiments}

\subsection{Datasets, Metrics, and Implementation}

\subsubsection{Datasets} 
We first evaluate methods on the well-known benchmark for monocular video reconstruction and NVS, DyCheck (iPhone)~\cite{gao2022monocular} . The datasets contain diverse casual strictly-monocular dynamic videos captured by a handheld camera, without circling the dynamic objects to approximate multi-view coverage. In addition, seven scenes include additional cameras that provide videos suitable for NVS evaluation. From these scenes, we select five representative dynamic scenes (Apple, Block, Teddy, Spin, and Paper-windmill), where Apple, Block, and Teddy feature hand-object interaction (HOI), Spin has a spinning human, and Paper-windmill is self-rotating.

\begin{figure}[h]
    \centering
    \begin{subfigure}[b]{0.26\textwidth}
        \centering
        \includegraphics[width=\textwidth]{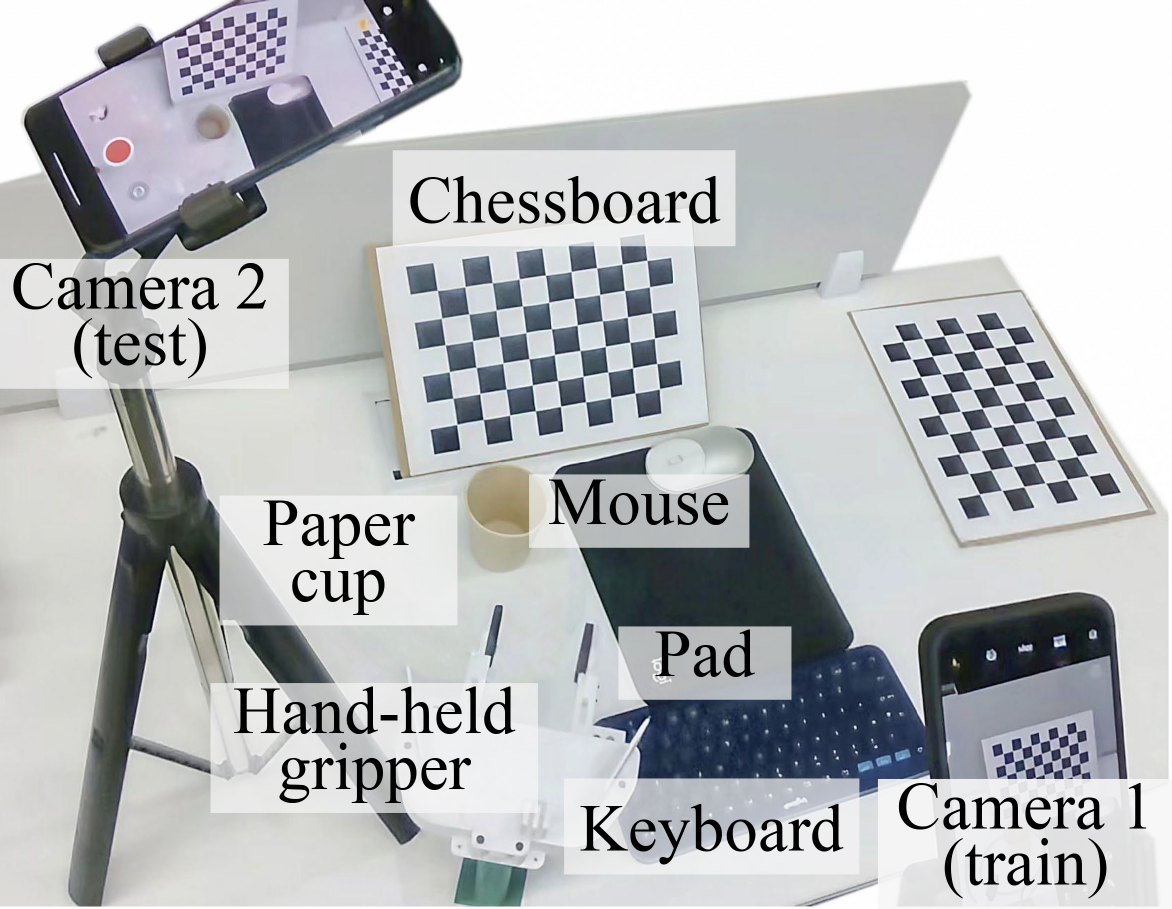}  
        \\[-1mm] 
        \caption{Setup.}
    \end{subfigure}
    \hspace{-2mm} 
    \begin{subfigure}[b]{0.22\textwidth}
        \centering
        \includegraphics[width=\textwidth]{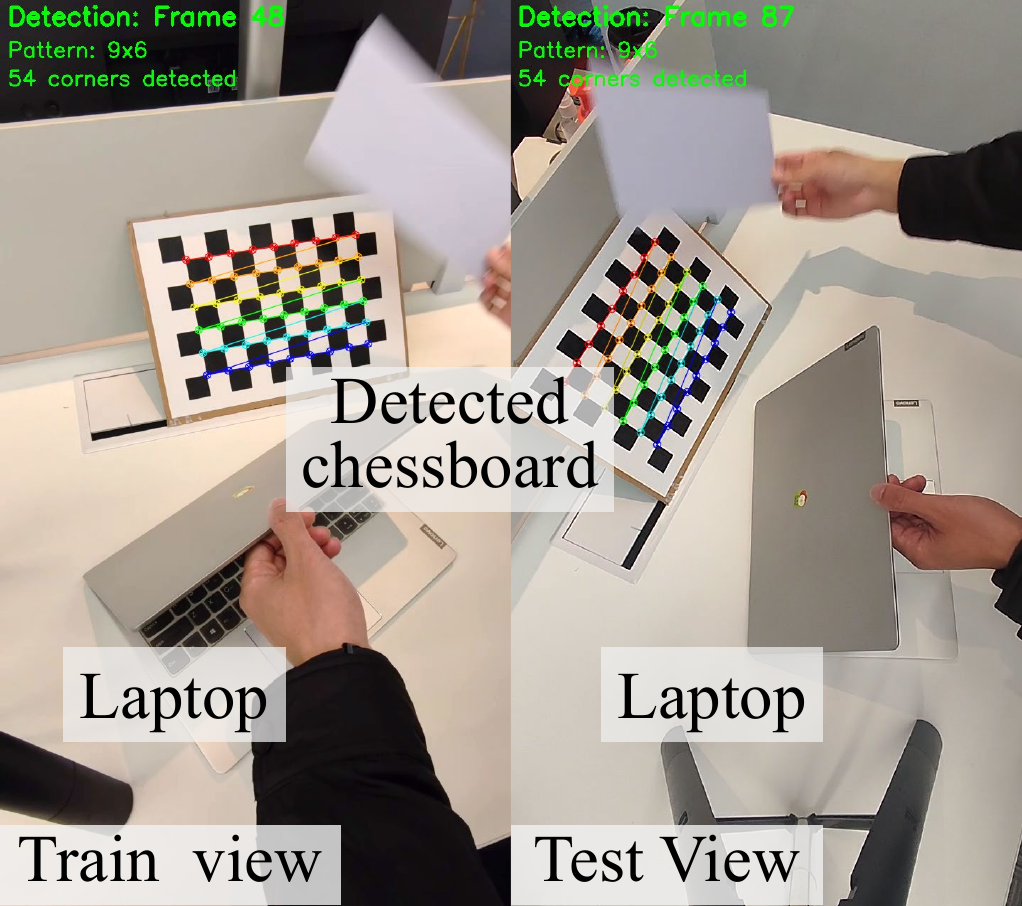}  
        \\[-1mm] 
        \caption{Time alignment.}
    \end{subfigure}
    \caption{Experimental setup for our custom datasets.}
    \label{fig:setup}
\end{figure}

We further evaluate our method on custom datasets, as shown in Figure~\ref{fig:setup}. Videos of dynamic scenes are also recorded from a strictly monocular viewpoint, with an additional fixed camera for evaluation. The videos of two cameras are temporally synchronized by chessboard detection, achieving frame-level alignment with greater stability than audio-based synchronization in~\cite{gao2022monocular}. Relative poses of cameras are calibrated using chessboards. Our datasets cover a diverse set of object categories, containing rigid objects (Keyboard), articulated objects (Laptop), and deformable objects (Paper-cup and Mouse-pad). These objects are manipulated by hands or a gripper.

\begin{table*}[t]
\centering
\vspace{2.5mm}
\caption{Evaluation of dynamic NVS on iPhone datasets~\cite{gao2022monocular}. Our method outperforms the sevral representative baselines, including NeRF-based dynamic reconstruction methods HyperNeRF~\cite{park2021hypernerf} and T-NeRF~\cite{gao2022monocular}, as well as dynamic Gaussian-based methods Deform-3DGS ~\cite{yang2024deformable}, Dynamic marbles~\cite{stearns2024dynamic}, and Shape-of-motion~\cite{wang2025shape}. (HOI: hand-object interaction)}
\begin{tabular}{l|ccc|ccc|ccc}
\toprule
\multirow{2}{*}{\textbf{Methods}} 
& \multicolumn{3}{c|}{\textbf{Overall}} 
& \multicolumn{3}{c|}{\textbf{Apple (HOI)}} 
& \multicolumn{3}{c}{\textbf{Block (HOI)}} \\
& mPSNR $\uparrow$ & mSSIM $\uparrow$ & mLPIPS $\downarrow$ 
& mPSNR $\uparrow$ & mSSIM $\uparrow$ & mLPIPS $\downarrow$ 
& mPSNR $\uparrow$ & mSSIM $\uparrow$ & mLPIPS $\downarrow$ \\
\midrule

HyperNeRF~\cite{park2021hypernerf} & 14.28 & 0.38 & 0.51 & 16.12 & 0.43 & 0.53 & 14.05 & 0.47 & 0.59 \\
T-NeRF~\cite{gao2022monocular}     & 15.09 & 0.41 & 0.50 & 15.98 & 0.38 & 0.60 & 14.38 & 0.52 & 0.55\\
Deform-3DGS~\cite{yang2024deformable} & 10.77 & 0.27 & 0.69 & 10.82 & 0.27  & 0.73 & 10.04 & 0.29 & 0.72 \\
Dynamic marbles~\cite{stearns2024dynamic} & 15.95 & - & 0.45 & 16.39 & - & 0.55 & 16.10 & - & \textbf{0.37} \\ 
Shape-of-motion~\cite{wang2025shape} &16.68 & 0.64 & 0.41
 &16.41 & 0.74 & 0.54 & 16.34     & 0.65    & 0.45    \\
\textbf{Ours}                & \textbf{17.07} & \textbf{0.66} & \textbf{0.38} 
                             & \textbf{17.15} & \textbf{0.76} & \textbf{0.51} 
                             & \textbf{16.49} & \textbf{0.67} & 0.42 \\
\midrule

\multirow{2}{*}{\textbf{Methods}} 
& \multicolumn{3}{c|}{\textbf{Paper-Windmill (rotating)}} 
& \multicolumn{3}{c|}{\textbf{Spin (spinning human)}} 
& \multicolumn{3}{c}{\textbf{Teddy (HOI)}} \\
& mPSNR $\uparrow$ & mSSIM $\uparrow$ & mLPIPS $\downarrow$ 
& mPSNR $\uparrow$ & mSSIM $\uparrow$ & mLPIPS $\downarrow$ 
& mPSNR $\uparrow$ & mSSIM $\uparrow$ & mLPIPS $\downarrow$ \\
\midrule
HyperNeRF~\cite{park2021hypernerf} & 14.59  & 0.26 & 0.37 & 14.61 & 0.42 & 0.44 & 12.04 & 0.31 & 0.62 \\
T-NeRF~\cite{gao2022monocular}     & 15.63 & 0.35 & 0.27 & 15.98 & 0.45 & 0.50 & 13.48 & 0.34 & 0.57 \\
Deform-3DGS~\cite{yang2024deformable} & 10.32 & 0.24 & 0.61  & 12.26     & 0.31   & 0.71 & 10.42 & 0.25 & 0.68 \\
Dynamic marbles~\cite{stearns2024dynamic} &  16.17 & - & 0.45 & 17.45 & - & 0.43 & 13.65 & - & \textbf{0.46} \\
Shape-of-motion~\cite{wang2025shape} & 19.50 & 0.56 & 0.22 & 17.42 & 0.70 & 0.30 & 13.73     & 0.55    & 0.53      \\
\textbf{Ours}                & \textbf{19.83} & \textbf{0.57} & \textbf{0.20} 
                             & \textbf{17.85} & \textbf{0.72} & \textbf{0.29} 
                             & \textbf{14.05} & \textbf{0.57} & 0.50 \\
\bottomrule
\end{tabular}
\label{tab:quanti-iphone}
\end{table*}

\begin{figure*}[t]
\centering
\includegraphics[width=0.98\textwidth]{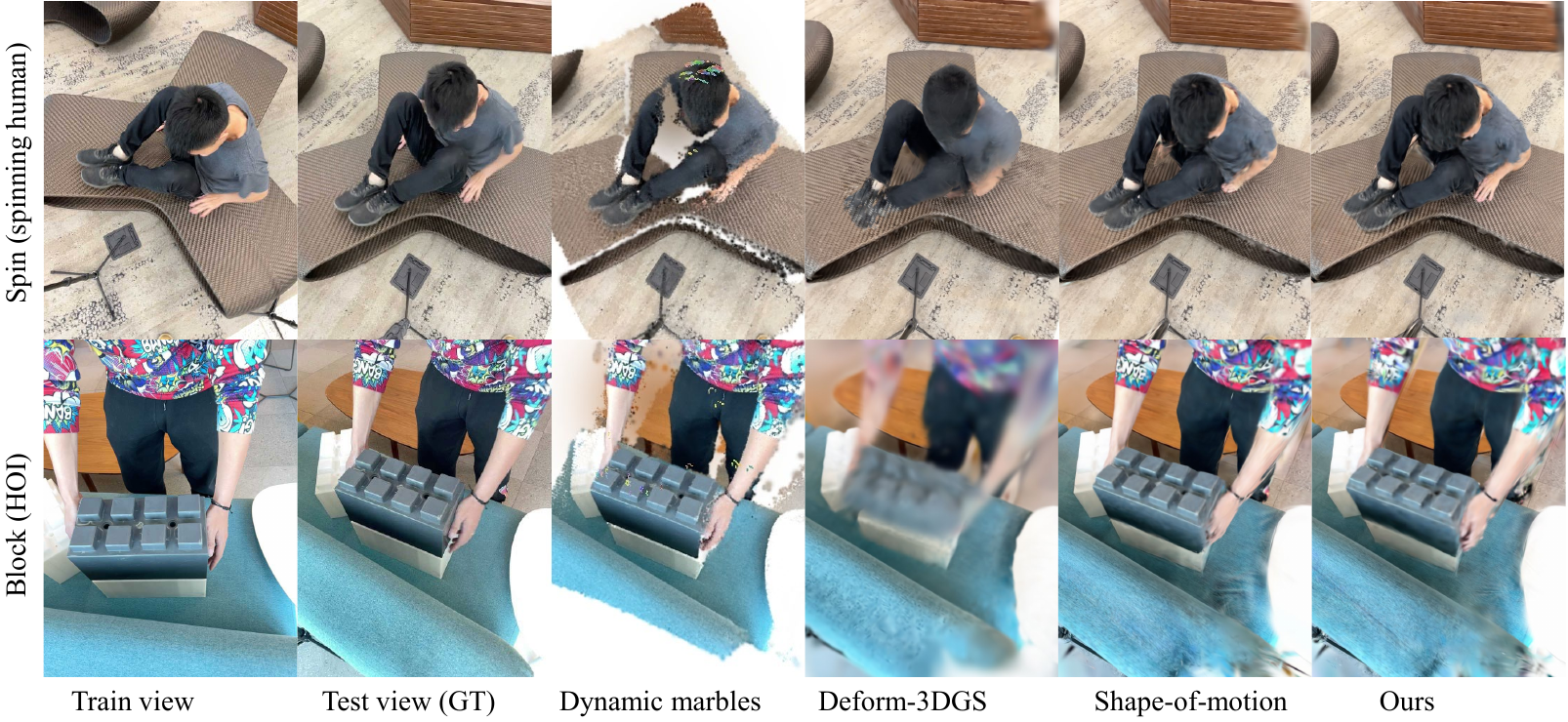} 
\caption{Qualitative results of NVS on iPhone datasets~\cite{gao2022monocular}. Ours synthesizes finer details than baselines. (GT: Ground truth)
}
\label{fig-qual-results}
\end{figure*}

\subsubsection{Metrics}
We adopt covisibility-masked image metrics~\cite{gao2022monocular}, \emph{e.g.}, mPSNR, mSSIM, and mLPIPS, for NVS evaluation since the test view includes regions that are not observed by the monocular training view. 

\subsubsection{Implementation} 
We preprocess monocular videos with off-the-shelf vision foundation models to acquire masks~\cite{yang2023track}, depths~\cite{Yang2024DepthAU, piccinelli2024unidepth}, tracks~\cite{Doersch2023TAPIRTA}, and camera poses~\cite{li2025megasam}.  
The canonical frame is selected as the one with the least occlusion, determined by cross-frame point-tracking visibility.
Adaptive dense control of both dynamic and static Gaussians follows the implementation of 3DGS~\cite{kerbl20233d}. 

For the MS-Dynamics configuration, we set $K_{\text{object}}=number \, of \, instances$ at the object level ($L_1$), $K_{\text{primitive}}=5$ at the sparse-primitive level ($L_2$), and $K_{\text{grain}}=10$ at the fine-grained level ($L_3$). This configuration is set according to the trade-off between motion expressiveness and optimization performance (see the ablation study Sec.~\ref{subsec-ablation}).

We adopt Adam~\cite{kinga2015method} for optimization and gsplat~\cite{ye2025gsplat} for CUDA-accelerated differentiable rasterization. Our method is implemented with PyTorch and trained with 500 epochs on a NVIDIA A800 GPU. Training a sequence containing 150 frames takes approximately 60 minutes. The rendering speed at test time reaches 40fps, allowing for real-time rendering.

\subsection{Results}

\begin{table*}[t]
\centering
\caption{Evaluation of dynamic NVS on custom datasets covering a wide range of object types with rigid, articulated, and deformable objects. Our method surpasses Shape-of-motion (state-of-the-art)~\cite{wang2025shape} across all scenes. }
\setlength{\tabcolsep}{1pt}
\begin{tabular}{l|ccc|ccc|ccc|ccc}
\toprule
\multirow{2}{*}{\textbf{Methods}} 
& \multicolumn{3}{c|}{\textbf{Keyboard (rigid)}} 
& \multicolumn{3}{c|}{\textbf{Laptop (articulated)}} 
& \multicolumn{3}{c|}{\textbf{Paper-cup (deformable)}}
& \multicolumn{3}{c}{\textbf{Mouse-pad (rigid-deformable)}}\\
& mPSNR$\uparrow$ & mSSIM$\uparrow$ & mLPIPS$\downarrow$ 
& mPSNR$\uparrow$ & mSSIM$\uparrow$ & mLPIPS$\downarrow$ 
& mPSNR$\uparrow$ & mSSIM$\uparrow$ & mLPIPS$\downarrow$ 
& mPSNR$\uparrow$ & mSSIM$\uparrow$ & mLPIPS$\downarrow$ \\
\midrule

Shape-of-motion~\cite{wang2025shape}  & 11.86  & 0.23 & 0.11 
                                       & 13.87  & 0.59 & 0.12  
                                       & 12.76  & 0.25  & 0.03
                                       &  8.41   & 0.47 & 0.13 \\
\textbf{Ours}                & \textbf{14.23} & \textbf{0.57} & \textbf{0.06} 
                             & \textbf{15.77} & \textbf{0.75} & \textbf{0.07} 
                             & \textbf{19.11} & \textbf{0.73} & \textbf{0.02} 
                             & \textbf{10.81} & \textbf{0.63} & \textbf{0.11} \\
\bottomrule
\end{tabular}
\label{tab:quanti-custom}
\end{table*}

\begin{figure*}[htbp]
    \centering
    \vspace{-2mm}
    \begin{subfigure}[b]{0.33\textwidth}
        \centering
        \includegraphics[width=\textwidth]{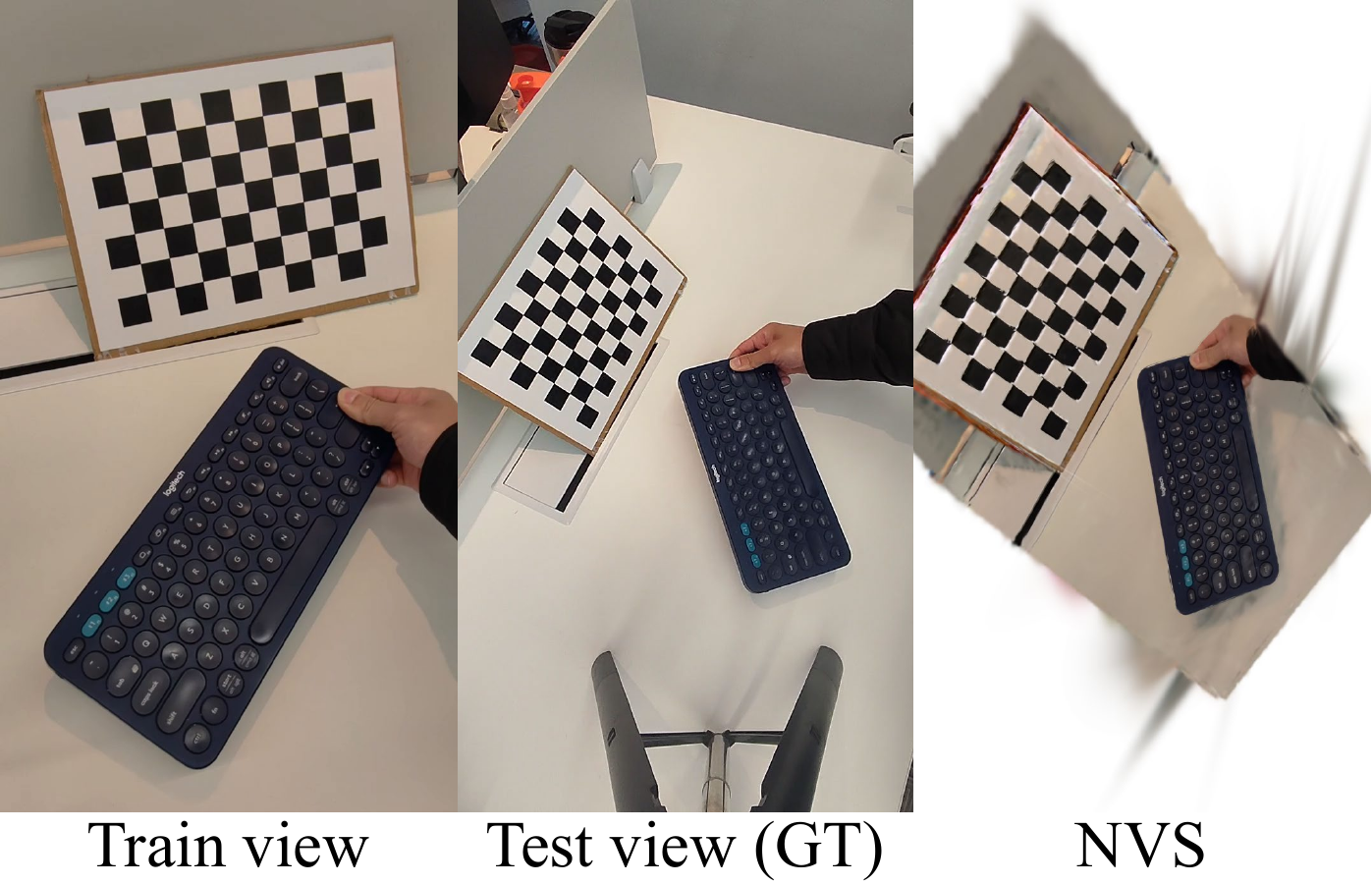}  
        \caption{Keyboard (rigid).}
    \end{subfigure}
    \hspace{-1.5mm}
    \begin{subfigure}[b]{0.33\textwidth}
        \centering
        \includegraphics[width=\textwidth]{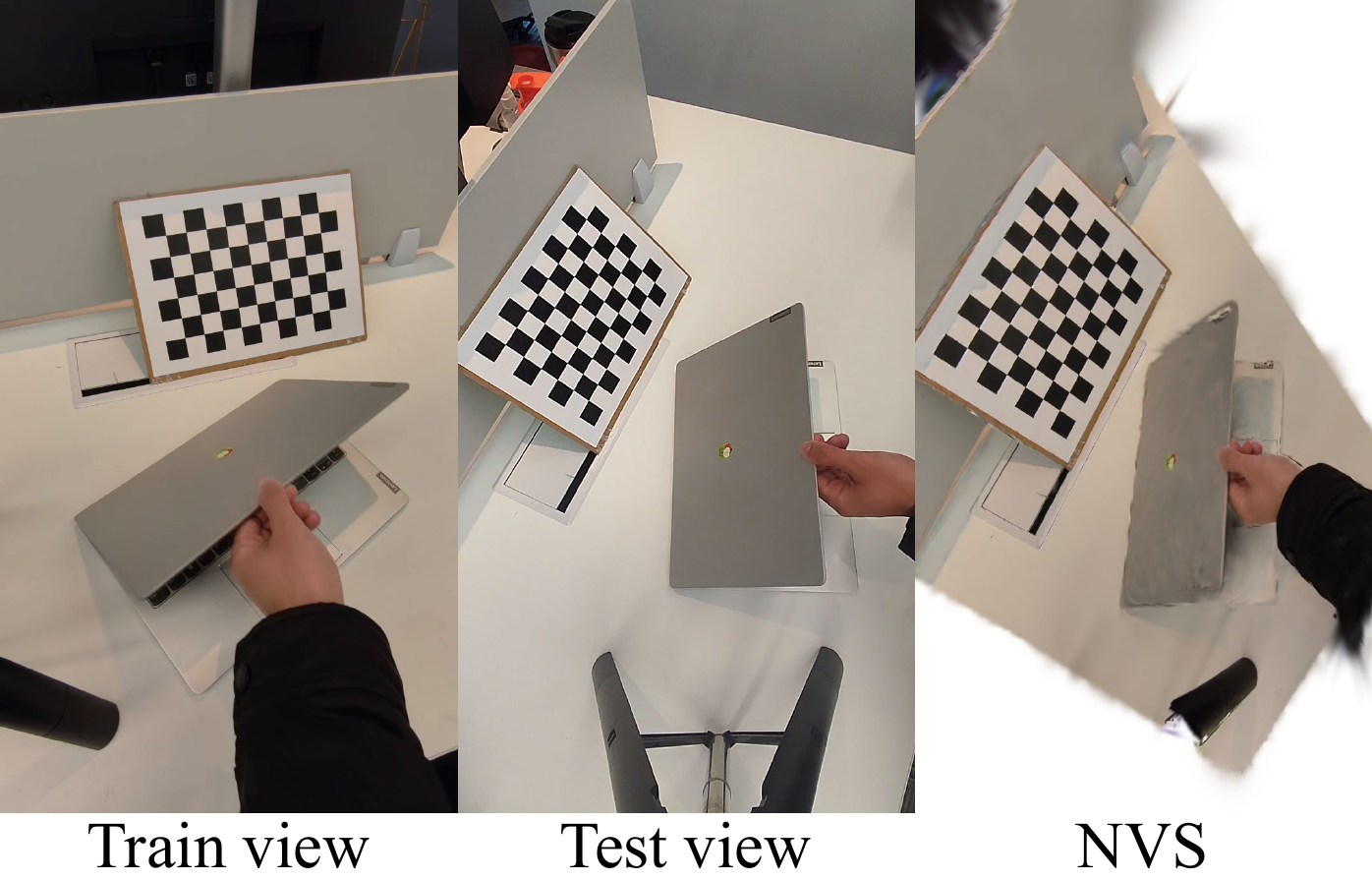}  
        \caption{Laptop (articulated).}
    \end{subfigure}
    \hspace{-1.5mm}
    \begin{subfigure}[b]{0.33\textwidth}
        \centering
        \includegraphics[width=\textwidth]{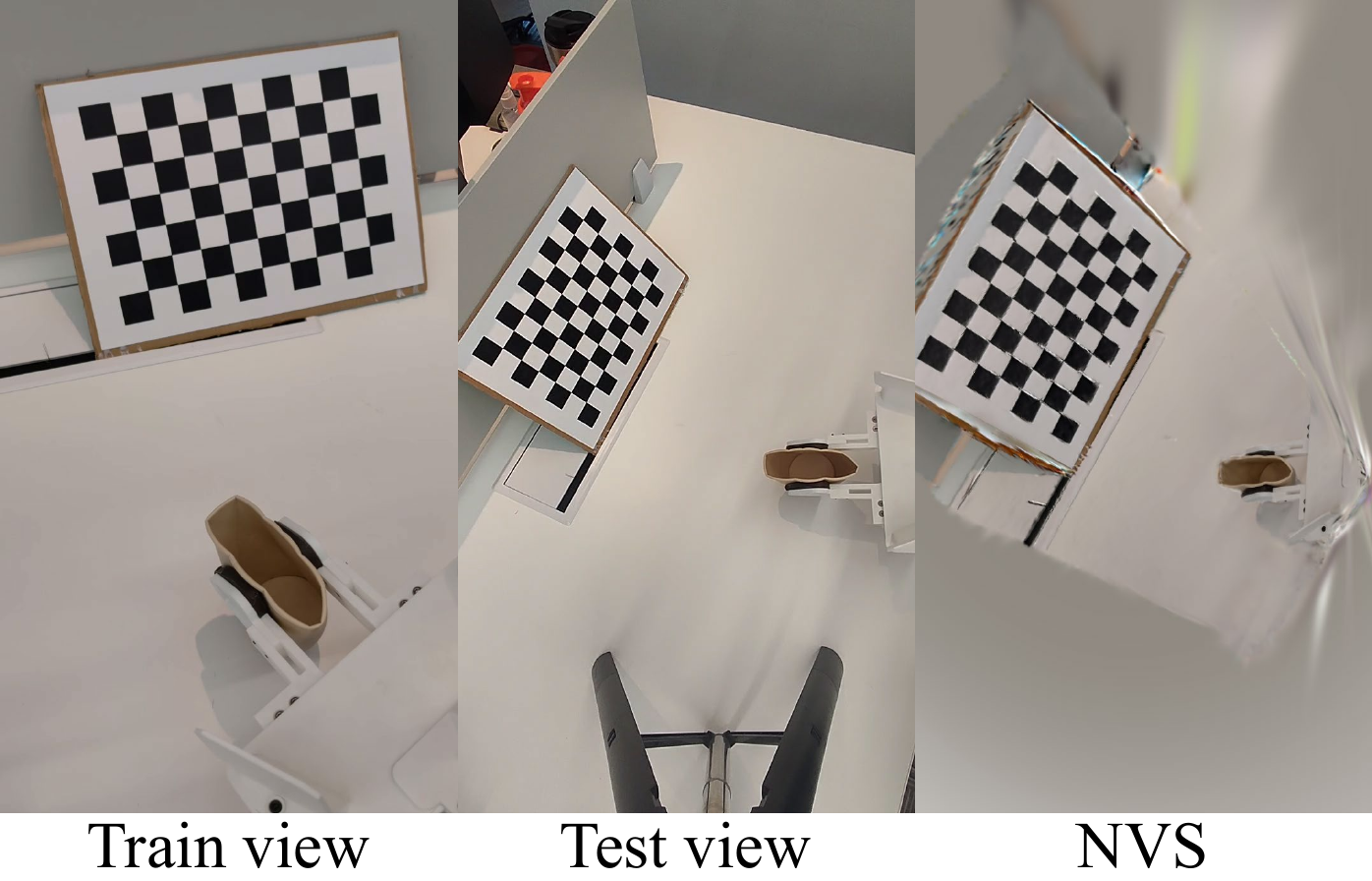}  
        \caption{Paper cup (deformable).}
    \end{subfigure}
    \caption{Qualitative results of dynamic NVS of our method on custom datasets containing hand- or gripper-object interactions. Our MS-Dynamics effectively represents these interaction dynamics and, even when trained under strictly monocular views, produces detailed novel views under considerable viewpoint changes.}
    \label{fig:custom}
\end{figure*}

We compare our method with several representative baselines. These include the NeRF-based dynamic reconstruction methods HyperNeRF~\cite{park2021hypernerf} and T-NeRF~\cite{gao2022monocular}, as well as dynamic Gaussian-based methods Deform-3DGS ~\cite{yang2024deformable}, Dynamic marbles~\cite{stearns2024dynamic}, and Shape-of-motion~\cite{wang2025shape}. Among them, T-NeRF and HyperNeRF extend neural radiance fields to dynamic settings. Deform-3DGS introduces MLP-based deformation fields into 3D Gaussian Splatting, while Dynamic marbles build isotropic Gaussian reconstructions from casual videos. Shape-of-motion~\cite{wang2025shape} explicitly models the motion field for Gaussians, which is a state-of-the-art method for dynamic monocular Gaussian reconstruction. These methods serve as strong baselines to evaluate the ability of our framework to handle dynamic monocular reconstruction.

Table~\ref{tab:quanti-iphone} presents quantitative results in iPhone datasets. Overall, our method achieves the best performance across all evaluation metrics with mPSNR of 17.07, mSSIM of 0.66, and mLPIPS of 0.38, consistently outperforming baselines. This is because T-NeRF and HyperNeRF rely heavily on an effective multi-view setting and struggle to handle strict monocular input. 
Deform-3DGS implicitly encodes Gaussian dynamics through a deformation field, which tends to oversmooth dynamic details and leads to poor reconstruction under monocular conditions. 
Dynamic marbles adopts isotropic Gaussian spheres for reconstruction, which reduces the degrees of freedom of Gaussian parameters in optimization but compromises the quality of NVS. 
Our method outperforms Shape-of-motion thanks to the proposed MS-Dynamics mechanism, enabling finer and more faithfully captured dynamic details.

Figure~\ref{fig-qual-results} presents qualitative results of dynamic NVS in iPhone datasets. Our method produces noticeably higher-quality renderings than the baselines, particularly in fine hand details. We primarily present rendering results from Gaussian-based methods whose rendering efficiency and fidelity are known to surpass NeRF-based methods~\cite{kerbl20233d}. 

Tab.~\ref{tab:quanti-custom} illustrates quantitative results in our custom datasets. We specifically compare our method with the state-of-the-art, Shape-of-motion. As shown, our approach outperforms existing methods across different object types with rigid, articulated, and deformable objects. This improvement is attributed to our effective decomposition of multi-scale interaction dynamics that guides the high-quality dynamic Gaussian reconstruction.

\begin{figure}
	\centering
	\includegraphics[width=.61\columnwidth]{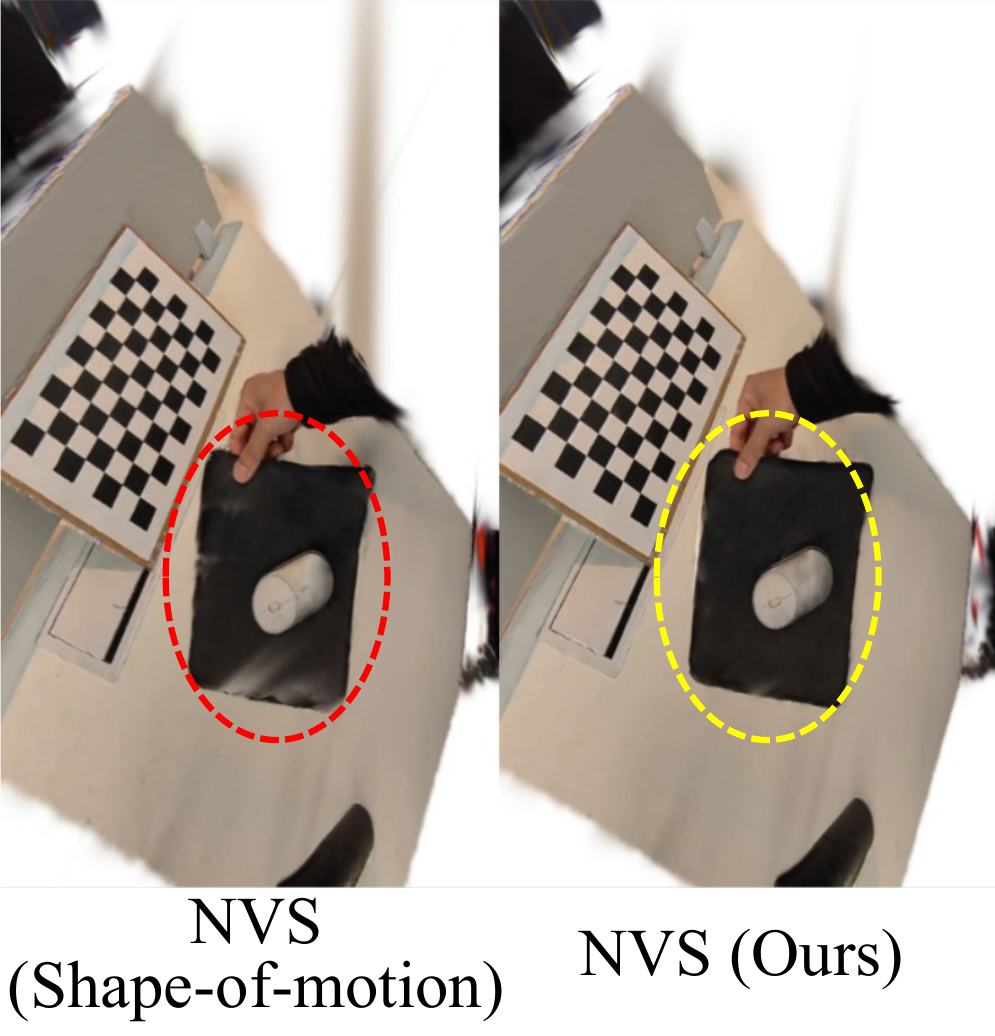}
	\caption{Qualitative results of dynamic NVS on the challenged custom scene Mouse-pad (rigid-deformable). Our method (yellow) achieves clearer novel-view-synthesis than the baseline Shape-of-motion (red). Please watch the supplementary video for more details.}
	\label{fig:custom4}
\end{figure}

Figures~\ref{fig:custom} and~\ref{fig:custom4} show the qualitative results of our methods. Under strictly monocular settings, MS-Dynamics effectively handles dynamic scenes with hand- or gripper-object interactions and successfully reconstructs high-fidelity novel views with significant viewpoint changes.

\subsection{Ablation}
\label{subsec-ablation}

\begin{figure*}[t]
    \centering
    \begin{subfigure}[b]{0.48\textwidth}
        \centering
        \includegraphics[width=\textwidth]{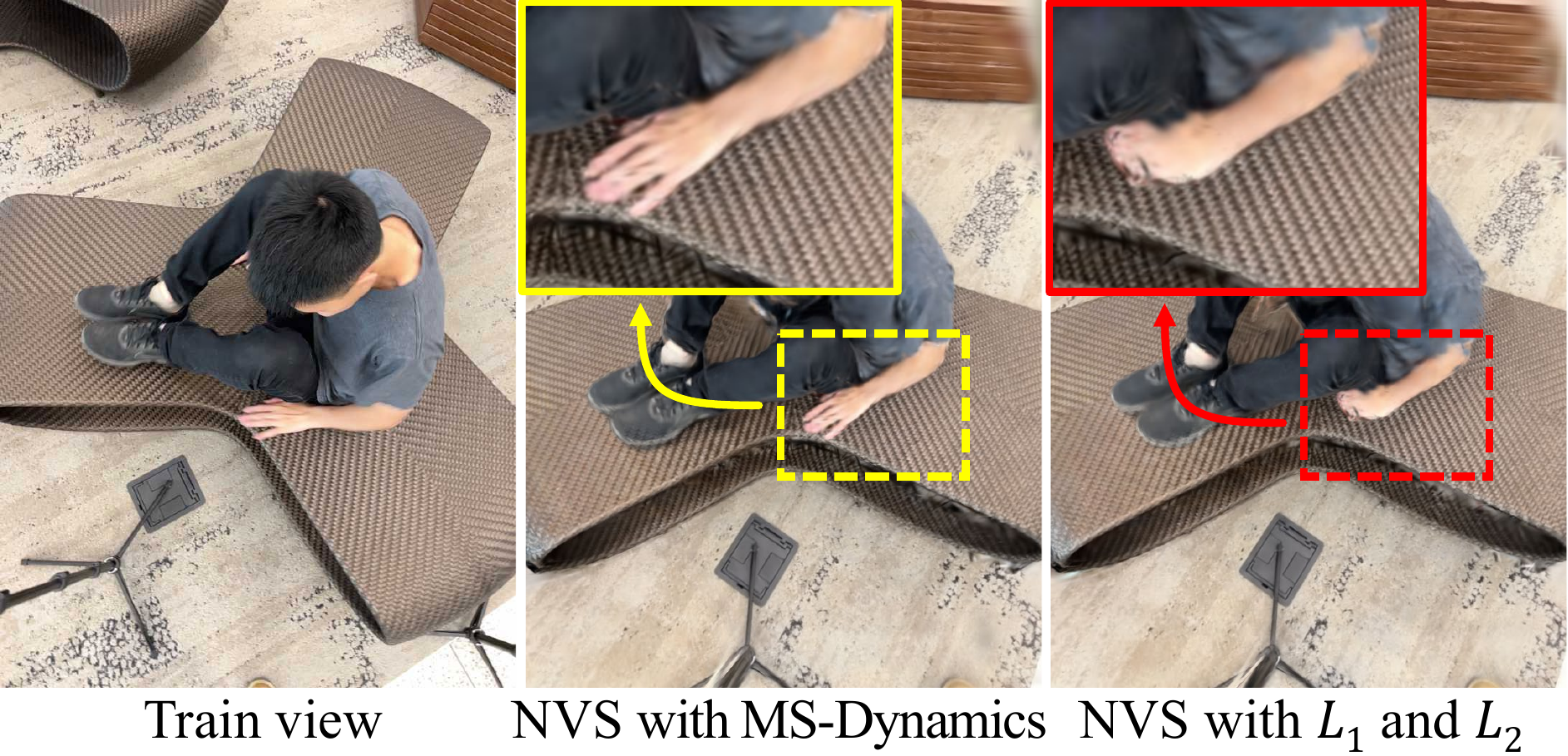}  
        \caption{Ablation of MS-Dynamics.}
        \label{fig:ablation on dynamics}
    \end{subfigure}
    \begin{subfigure}[b]{0.48\textwidth}
        \centering
        \includegraphics[width=\textwidth]{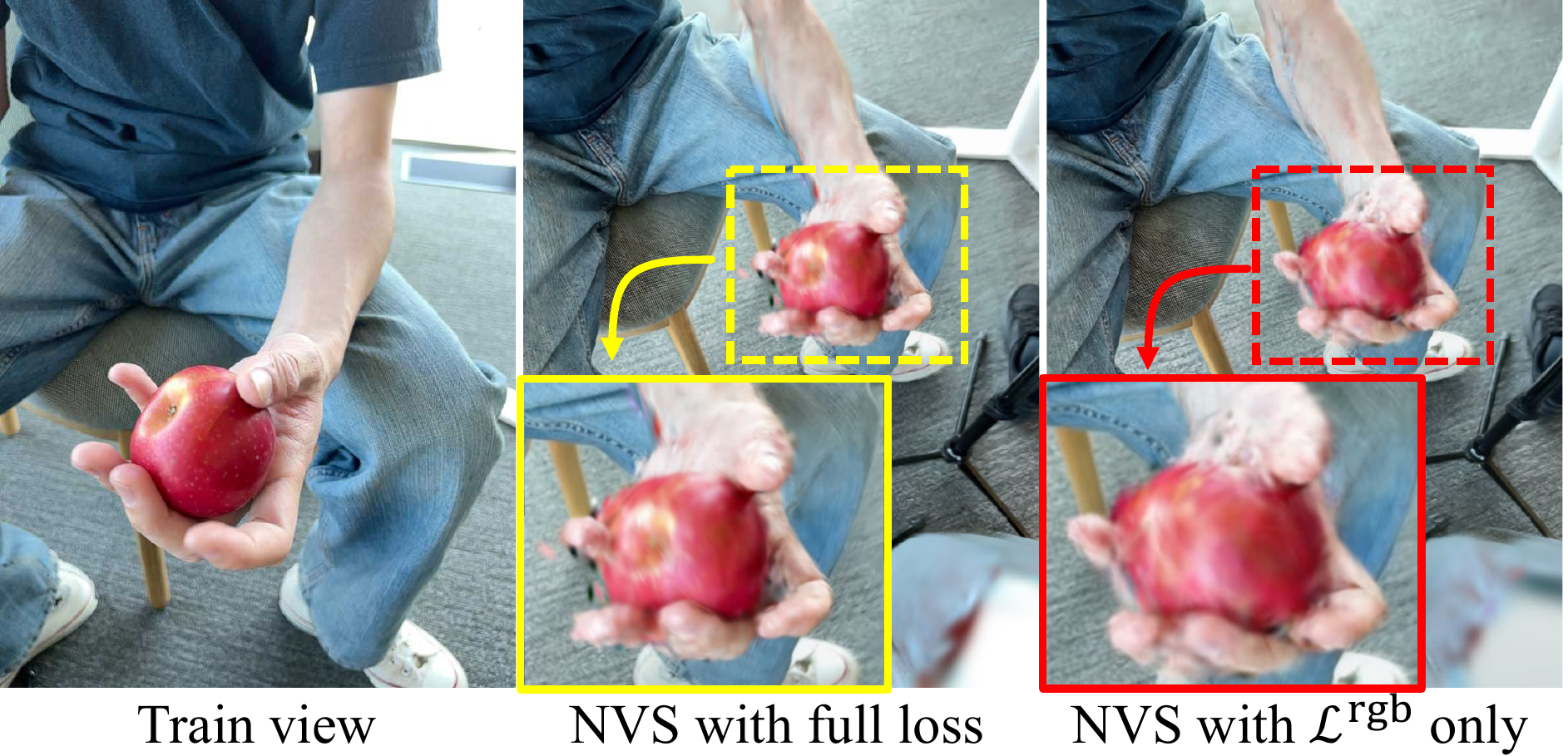}  
        \caption{Ablation of loss function.}
        \label{fig:ablation on loss}
    \end{subfigure}
    \caption{Qualitative results of ablation studies in iPhone datasets~\cite{gao2022monocular}. Compared to our method, the variant with motion representation of $L_1$--$L_2$ (a) or one with only RGB-loss $\mathcal{L}^{\text {rgb}}$ (b) yields blurrier NVS results.}
    \label{fig:ablation}
\end{figure*}

As shown in Table~\ref{tab:ablation}, the ablation study evaluates how different representations of dynamics and loss configurations affect reconstruction quality on the iPhone datasets~\cite{gao2022monocular}. The results demonstrate that progressively enriching the motion representation---from object-level dynamics ($L_1$ only, mPSNR = 11.34) to incorporating sparse motion primitives at the primitive level ($L_1$ and $L_2$, mPSNR = 16.70)---yields substantial improvements across all metrics, highlighting the importance of fine-grained motion modeling. 

Interestingly, further increasing the density of $L_2$ primitives (e.g., raising $K_{\text {primitive }}$ from 5 to 10, denoted as $L_2+$) leads to a slight performance drop (mPSNR = 16.02), suggesting a trade-off between motion expressiveness and NVS performance when over-parameterizing intermediate motion layers.  
This indicates that deepening the motion hierarchy by introducing fine-grained residual modes at $L_3$ is more effective than simply expanding the breadth of intermediate primitives, as these residual components further enhance the expressiveness of the motion field.

We also experimented with an additional dense particle-level layer ($L_1$--$L_4$), but observed no significant improvement in reconstruction quality (mPSNR = 16.81 vs.\ 17.07) while incurring substantially longer optimization time. This suggests that the motion representation capacity of the $L_1$--$L_3$ hierarchy is already sufficient to capture the dominant deformation modes, and further refinement leads to diminishing returns due to motion-scale saturation.

Moreover, replacing the full loss with RGB-only supervision ($\mathcal{L}^{\text{rgb}}$ only) achieves competitive results (mPSNR = 16.60), yet still lags behind the complete formulation. Only when combining the full MS-Dynamics hierarchy ($L_1$--$L_3$) with the foundation-prior regularized loss (Eq.~\ref{eq-loss}) do we obtain the best overall performance (mPSNR = 17.07, mSSIM = 0.66, mLPIPS = 0.38). This confirms that both components, including structured multi-scale dynamics and foundation-prior loss regularization, are essential for high-fidelity dynamic novel view synthesis.

\begin{table}[h]
\centering
\caption{Ablation of MS-Dynamics and loss functions on Dycheck-iPhone~\cite{gao2022monocular}. Without MS-Dynamics or foundation-prior loss, the performance of dynamic NVS is degraded. (Variants of MS-Dynamics, $L_1$ only: object-level motion patterns; $L_1$ and $L_2$: object- and sparse-primitive-level motion representation; $L_1$ and $L_2+$: $L_1$ with denser $L_2$ motion primitives ($K_{\text{primitive}} = 10$ vs.\ baseline $K_{\text{primitive}} = 5$). Variants of loss function, $\mathcal{L}^{\text {rgb}}$ only: supervision with only RGB loss.) }

\setlength{\tabcolsep}{4.0pt}   
\begin{tabular}{l|ccc}
\toprule
\multirow{2}{*}{\textbf{Variant methods}}  & \multicolumn{3}{c}{\textbf{iPhone datasets}~\cite{gao2022monocular}} \\
& mPSNR $\uparrow$ & mSSIM $\uparrow$ & mLPIPS $\downarrow$ \\
\midrule
$L_1$ only               &  11.34    & 0.30   &  0.67    \\
$L_1$ and $L_2$                  & 16.70  & 0.63 & 0.41   \\
$L_1$ and $L_2 +$                & 16.02  & 0.61 & 0.43      \\
\midrule
$\mathcal{L}^{\text {rgb}}$ only      & 16.60      &0.63  &   0.39  \\
\midrule

\makecell[l]{MS-Dynamics ($L_1$--$L_3$) \\ and full loss (Eq. (\ref{eq-loss}))}
    & \textbf{17.07} & \textbf{0.66} & \textbf{0.38} \\

\bottomrule
\end{tabular}
\label{tab:ablation}
\end{table}

Figure~\ref{fig:ablation} shows some qualitative results of ablation. It can be observed that our method with MS-Dynamics (Figure~\ref{fig:ablation on dynamics}) and full loss (Figure~\ref{fig:ablation on loss}) produces clearer NVS results, whereas the variant with motion representation of $L_1$--$L_2$ or one with only RGB-loss yields blurrier reconstructions. This further demonstrates the importance of the MS-Dynamics mechanism and the foundation-prior loss.

\section{Conclusion}

In this work, we propose Gaussian sequences with MS-Dynamics as a unified framework for 4D reconstruction from monocular videos. By factorizing complex motion into structured levels and integrating complementary multi-modal cues of vision foundation, our method effectively reduces reconstruction ambiguity and stabilizes optimization. Experiments on benchmark and real-world custom data demonstrate clear gains in fidelity of dynamic NVS. 
Despite these achievements, our method may still struggle with severe occlusions or large-scale topology changes under strictly monocular conditions; extending the framework toward more robust representation of dynamics and applying it to generation of robot learning data are promising directions for future work.









 \bibliographystyle{ieeetr}
 \bibliography{references}
\end{document}